\documentclass[journal]{IEEEtran}
\usepackage{amsmath,amsfonts}
\usepackage{algorithmic}
\usepackage{algorithm}
\usepackage{array}
\usepackage{textcomp}
\usepackage{stfloats}
\usepackage{url}
\usepackage{verbatim}
\usepackage{graphicx}
\usepackage{balance}
\usepackage{multirow}

\begin{document}

\IEEEpubid{%
\begin{minipage}{\textwidth}
\centering\footnotesize
This work has been submitted to the IEEE for possible publication.\\
Copyright may be transferred without notice, after which this version may no longer be accessible.
\end{minipage}%
}

\title{Toward Efficient Spiking Transformers: Synapse Pruning Meets Synergistic Learning-Based Compensation}

\author{Hongze Sun,	Wuque Cai, Duo Chen, Quan Tang, Shifeng Mao, Jiayi He, Zhenxing Wang, Yan Cui, Dezhong Yao, \emph{Senior Member, IEEE}, Daqing Guo
	\thanks{This work was supported in part by the National	Key Research and Development Program of China (2023YFF1204200), in part by the STI 2030–Major Projects (2022ZD0208500), and in part by the Sichuan Science and Technology Program (2024NSFJQ0004, 2024NSFTD0032, and DQ202410). (Corresponding authors: Dezhong Yao; Daqing Guo.)
		
		Hongze Sun, Wuque Cai, Duo Chen, Quan Tang, Shifeng Mao, Jiayi He, Zhenxing Wang, Daqing Guo are with Clinical Hospital of Chengdu Brain Science Institute, MOE-K Lab for NeuroInformation, Int Institute of Brain-Apparatus Communication, University of Electronic Science and Technology of China, Chengdu, China, and also with School of Life Science and Technology, University of Electronic Science and Technology of China, Chengdu, China (e-mail: dqguo@uestc.edu.cn).
		
		Yan Cui is with MOE-K Lab for NeuroInformation, Sichuan Academy of Medical Sciences and Sichuan Provincial People’s Hospital, University of Electronic Science and Technology of China, Chengdu, China.
		
		Dezhong Yao is with Clinical Hospital of Chengdu Brain Science Institute, MOE-K Lab for NeuroInformation, Int Institute of Brain-Apparatus Communication, University of Electronic Science and Technology of China, Chengdu, China, also with School of Life Science and Technology, University of Electronic Science and Technology of China, Chengdu, China, also with the Research Unit of NeuroInformation (2019RU035), Chinese Academy of Medical Sciences, Chengdu 611731, China, and also with the School of Electrical Engineering, Zhengzhou University, Zhengzhou 450001, China (e-mail: dyao@uestc.edu.cn).
	}
\\ 
}

\maketitle

\begin{abstract}
	As a foundational architecture of artificial intelligence models, Transformer has been recently adapted to spiking neural networks with promising performance across various tasks. However, existing spiking Transformer~(ST)-based models require a substantial number of parameters and incur high computational costs, thus limiting their deployment in resource-constrained environments. To address these challenges, we propose combining synapse pruning with a synergistic learning-based compensation strategy to derive lightweight ST-based models. Specifically, two types of tailored pruning strategies are introduced to reduce redundancy in the weight matrices of ST blocks: an unstructured $\mathrm{L_{1}P}$ method to induce sparse representations, and a structured DSP method to induce low-rank representations. In addition, we propose an enhanced spiking neuron model, termed the synergistic leaky integrate-and-fire (sLIF) neuron, to effectively compensate for model pruning through synergistic learning between synaptic and intrinsic plasticity mechanisms. Extensive experiments on benchmark datasets demonstrate that the proposed methods significantly reduce model size and computational overhead while maintaining competitive performance. These results validate the effectiveness of the proposed pruning and compensation strategies in constructing efficient and high-performing ST-based models.
\end{abstract}


\begin{figure}[!t]
	\centering
	\includegraphics[width=1.0\columnwidth]{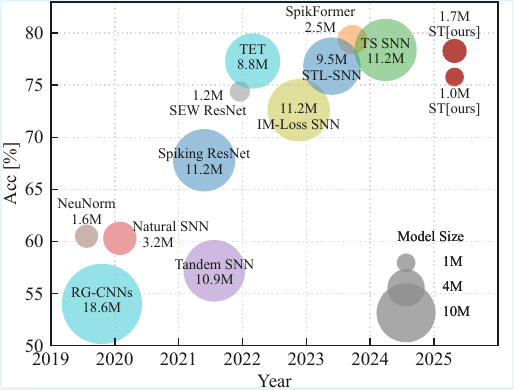}
	\caption{Comparison of model parameters and classification accuracy between our lightweight models (denoted as `ST[ours]') and existing ST-based models on the CIFAR10-DVS dataset.}
	\label{Fig1}
\end{figure}

\section{Introduction}\label{Introduction}
\IEEEPARstart{T}{hE} Transformer architecture~\cite{vaswani2017attention} has emerged as a foundational backbone for a wide array of large language models~\cite{achiam2023gpt,touvron2023llama}, owing to its strengths in modeling long-range dependencies, adaptability across multiple modalities, and high parallelism. Recently, researchers have extended Transformer architectures to spiking neural networks (SNNs), giving rise to spiking Transformer (ST)-based models~\cite{zhou2023spikformer,yao2023spike}. By making targeted adaptations to the dynamics of spiking neurons, these models have demonstrated significant improvements across a variety of complex tasks~\cite{yao2024spikedriven,zhou2024qkformer,10772601}. However, the ST is inherently a parameter-intensive architecture, where performance often correlates with the model size~\cite{dosovitskiy2021an,pnas2528527122}, resulting in substantial computational and efficiency costs. Achieving parameter-efficient adaptations and low computational cost thus remains a critical challenge in this field.

\enlargethispage{-36pt}

To enhance the efficiency of ST blocks, numerous methods have been proposed in prior work. Among these, structure-oriented optimization represents a key direction, focusing on architectural modifications such as spatio-temporal pruning~\cite{zhou2024spatial}, meta-architecture redesign~\cite{yao2024spikedriven}, and the integration of lightweight functional modules~\cite{zhou2024qkformer}. These techniques aim to improve representational efficiency within ST blocks. Additionally, engineering-oriented approaches—such as quantization-aware training~\cite{qiu2025quantized} and specialized neuromorphic processors~\cite{kim2025c}—address practical concerns like energy consumption and inference latency, facilitating the deployment of ST-based models on resource-constrained edge devices. Despite their effectiveness, these approaches are often tightly coupled with specific model architectures or constrained by task-specific assumptions, limiting their generalizability. In contrast, our goal is to develop a universal, model-agnostic compression strategy for ST blocks without compromising performance.

SNNs are composed of biologically inspired spiking neurons that emulate the spatio-temporal dynamics observed in human brain activity. Drawing inspiration from these biological mechanisms, researchers have proposed a range of enhanced spiking neuron models~\cite{qiu2025quantized,huang2024towards,wang2023complex,fang2021incorporating}. Notably, models such as IE-LIF~\cite{qiu2025quantized}, multi-threshold~\cite{huang2024towards}, and dynamic-threshold neurons~\cite{wang2023complex} have been integrated into ST blocks to improve the efficiency of spiking units. In this work, we aim to propose an enhanced spiking neuron model that is efficient and simple to implement and train. We hope that the enhanced spiking neuron model can improve the efficiency of ST blocks, leading to more lightweight ST-based models. 

To tackle the aforementioned challenges, we introduce a novel lightweight method that combines synapse pruning with a synergistic learning-based compensation strategy, aiming to construct lightweight yet high-performing ST-based models. We propose two types of customized pruning strategies tailored to the weight matrices within ST blocks: (1) an unstructured $\mathrm{L_{1}}$ norm-based parameter sparsification ($\mathrm{L_{1}P}$) method that promotes sparsity by removing weak synaptic connections, and (2) a structured dimension significance-based pruning (DSP) method that induces low-rank representations by reducing the dimensionality of patch embeddings. To mitigate the potential performance degradation caused by synapse pruning, we further introduce an enhanced spiking neuron model, termed the synergistic leaky integrate-and-fire (sLIF) neuron, which jointly leverages synaptic and intrinsic plasticity. Through synergistic learning between these two plasticity mechanisms, the sLIF neuron effectively compensates for the performance loss introduced by synapse pruning.

Our method addresses two key advantages. First, the pruning pipeline allows flexible control over model size by adjusting the desired sparsity level. Second, the proposed sLIF neuron is plug-and-play, enabling seamless integration into existing ST-based models. As shown in Fig.~\ref{Fig1}, our approach achieves higher accuracy with fewer parameters compared to existing ST-based models on the neuromorphic CIFAR10-DVS dataset. 

The main contributions and highlights of this study can be summarized as follows.
\begin{itemize}
	\item Two tailored pruning strategies are developed specifically for the ST block to enable efficient compression of ST-based models.
	\item An enhanced sLIF neuron model is proposed to effectively compensate for the information loss based on the synergistic learning between synaptic and intrinsic plasticity.
	\item Experiments across diverse tasks demonstrate that our method achieves significant model compression while maintaining competitive performance.
\end{itemize}

The remainder of this article is organized as follows. Section~\ref{Related Work} reviews prior studies on efficient spiking transformers and bio-inspired spiking neuron models. Section~\ref{Methods} introduces the proposed lightweight strategy for spiking transformers and details its compensation mechanism based on synergistic learning. Section~\ref{Experiments and Results} describes the experimental framework, including the setup, results, and corresponding analyses. Finally, Section~\ref{Discussion and Conclusion} summarizes the key findings and concludes the paper.

\thispagestyle{plain}
\markboth{}{}
\section{Related Work}\label{Related Work}
In this section, we briefly review recent works on efficient spiking transformers and bio-inspired spiking neuron models that are closely related to our study.

\subsection{Efficient Spiking Transformers}
ST-based models have garnered significant attention due to their potential for biologically plausible computation and energy-efficient processing of spatio-temporal data~\cite{zhou2023spikformer,yao2023spike}. However, the high computational cost and memory overhead inherent in their complex dynamics necessitate efficiency-oriented enhancements. Prior work can be broadly categorized into structure-oriented and engineering-oriented approaches. 

Structure-oriented optimization: A  primary direction for improving efficiency in ST blocks lies in architectural redesign. Inspired by model pruning strategies, researchers have explored eliminating redundant spatial and temporal components in attention mechanisms and feedforward layers, thereby reducing computational burden without substantially degrading accuracy~\cite{zhou2024spatial}. Another line of work proposes meta-architecture redesigns~\cite{yao2024spikedriven,wang2025spiking}, which tailor model structures to the unique characteristics of spiking data. Furthermore, the integration of lightweight functional modules—such as linearized QK-value embedding and simplified gating mechanisms—have shown promise in improving representational efficiency with minimal performance trade-offs~\cite{zhou2024qkformer}. 

Engineering-oriented solutions: In parallel, engineering-based methods addressing practical deployment issues have been proposed. Weight quantization, which reduces numerical precision, is a widely used technique to enable model deployment on edge devices. Researchers have incorporated quantization-aware training into ST-based models to lower the bit-width of weights and activations, thereby minimizing memory footprint and enabling deployment on low-power hardware~\cite{qiu2025quantized}. Similarly, advances in specialized neuromorphic hardware—such as event-driven computing cores and custom-designed crossbar arrays—have significantly reduced latency and energy consumption for ST-based models~\cite{kim2025c}. Despite their success, these methods tend to be model-dependent or require non-trivial system-level integration. This lack of universality limits their adaptability to diverse architectures and downstream tasks.

\subsection{Bio-Inspired Spiking Neuron Models}
The spiking neuron, as the fundamental computational unit, plays a critical role in determining the overall performance of SNNs~\cite{zheng2024temporal}. Among various neuron models, the LIF model is one of the most widely adopted due to its simplicity and computational efficiency. By training LIF neurons with synaptic plasticity mechanisms~\cite{wu2018spatio}, SNN models have achieved competitive performance on various tasks~\cite{sun2024reliable,10318216}. Despite these advantages, the limited representational capacity of LIF in capturing complex spatio-temporal dynamics has motivated the development of more biologically inspired neuron models. 

In spiking neurons, intrinsic parameters (i.e., the membrane time constant, firing threshold, and resting potential) play a crucial role in shaping internal dynamics, including input integration, spike generation, and the refractory period~\cite{9449951,klinshov2024extending}. Based on the biological mechanisms underlying the firing threshold, numerous enhanced variants of the LIF model have been proposed, focusing on the threshold adaptation, dynamic thresholds, and subthreshold dynamics~\cite{sun2023synapse,huang2024towards}. Compared to the standard LIF neuron, these models offer a more biologically plausible depiction of threshold behavior, thereby enhancing the neuronal ability to robustly represent spatial patterns. Furthermore, the membrane time constant governs the temporal integration of incoming signals~\cite{pazderka2024role}. By introducing heterogeneity in this parameter, recent studies have significantly improved the capacity of SNNs to capture temporal features across multiple time scales~\cite{fang2021incorporating,10857949}. 

A key challenge in enhancing LIF neuron models lies in the effective integration of the synergistic interactions between synaptic and intrinsic parameters. Prior studies have typically treated these parameters as independent components or optimized them using information-theoretic approaches, often overlooking their interdependence~\cite{li2013synergies,10235316}. This not only limits the model expressiveness but also complicates the training process. Therefore, two critical questions remain open: how to incorporate synaptic and intrinsic plasticity in a unified and efficient framework, and how to enable synergistic learning between these two forms of plasticity.

\begin{figure*}[!t]
	\centering
	\includegraphics[width=2.\columnwidth]{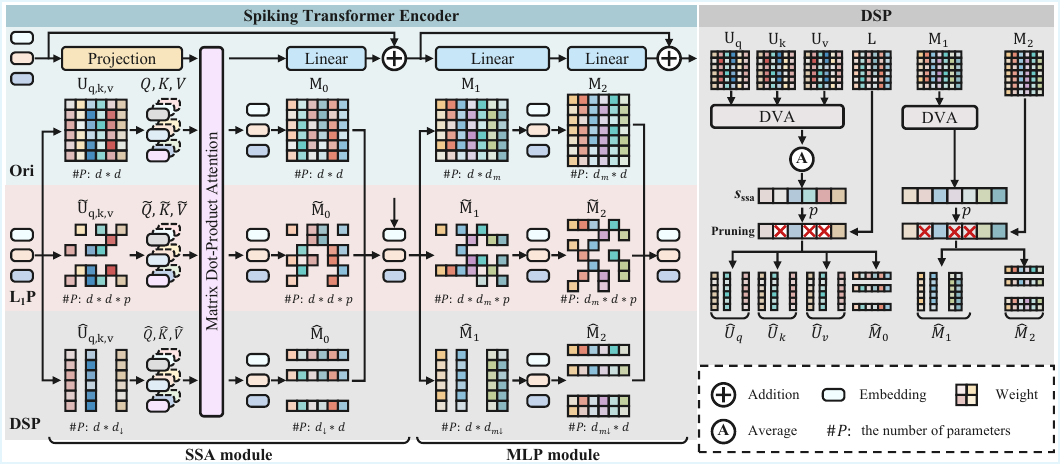}
	\caption{Overview of the proposed lightweight ST-based models and the corresponding pruning strategies. In the original ST-based encoder, input embeddings are sequentially processed by the SSA and MLP modules. The primary parameter overhead resides in matrices $\mathbf{U}{\mathrm{q}}$, $\mathbf{U}{\mathrm{k}}$, $\mathbf{U}{\mathrm{v}}$, and $\mathbf{M}{0}$ of the SSA module, as well as $\mathbf{M}{1}$ and $\mathbf{M}{2}$ of the MLP module. To address this, two lightweight strategies are introduced: L$_{1}$P, which yields sparse matrices, and DSP, which produces low-rank matrices.}
	\label{Fig2}
\end{figure*}

\section{Methods}\label{Methods}
In this section, we first propose two synapse pruning strategies tailored for spiking Transformers, encompassing both unstructured and structured approaches. Subsequently, we introduce an sLIF neuron model that concurrently incorporates synaptic and intrinsic plasticity, along with a compensation method based on synergistic learning.

\subsection{Efficient Compression for Spiking Transformer}
A standard Spiking Transformer~(ST) encoder can be formally described as follows:
\begin{align}
x^{\prime}_l &= \mathrm{SSA}(x_{l-1})+x_{l-1},  &&l = 1, \dots, L, \label{eq4}\\
x_l &= \mathrm{MLP}(x^{\prime}_l)+x^{\prime}_l,  &&l = 1, \dots, L.\label{eq5}
\end{align}
Here, the SSA and MLP represent the spiking self-attention module and linear layers module, respectively. In an ST-based model, the spatio-temporal patch embeddings $x_0 \in \mathbb{R}^{T \times N \times d}$ are progressively refined through $L$ stacked $\mathrm{ST}$ blocks. The resulting high-level representations are subsequently fed into a task-specific prediction head to produce the final output.

Specifically, in the SSA module, the input $x \in \mathbb{R}^{T \times N \times d}$, consisting of $N$ patch embeddings each of dimension $d$, is first linearly projected into query ($Q$), key ($K$), and value ($V$) representations using three learnable projection matrices $\mathbf{U}_{\mathrm{q}} \in \mathbb{R}^{d \times d}$, $\mathbf{U}_{\mathrm{k}} \in \mathbb{R}^{d \times d}$ and $\mathbf{U}_{\mathrm{v}} \in \mathbb{R}^{d \times d}$, respectively:
\begin{equation}
[Q,K,V] = [x\mathbf{U}_{\mathrm{q}},x\mathbf{U}_{\mathrm{k}},x\mathbf{U}_{\mathrm{v}}].
\label{eq6}
\end{equation}
The attention output is computed using a scaled dot-product attention mechanism followed by a linear transformation:
\begin{align}
x_{\mathrm{attn}} &= \frac{QK^{\mathrm{T}}}{\sqrt{d}}V, \\
\mathrm{SSA}(x) &= x_{\mathrm{attn}} \mathbf{M}_0,
\end{align}
where $\mathbf{M}_0 \in \mathbb{R}^{d \times d}$ denotes the weight matrix of the post-attention linear layer. Subsequently, in the MLP module, two linear layers with weight matrices $\mathbf{M}_1 \in \mathbb{R}^{d \times d_\mathrm{m}}$ and $\mathbf{M}_2 \in \mathbb{R}^{d_\mathrm{m} \times d}$ are employed to generate the final output of the ST block:
\begin{equation}
\mathrm{MLP}(x^{\prime}) = x^{\prime} \mathbf{M}_1 \mathbf{M}_2,
\end{equation}
where $d_\mathrm{m}$ represents the hidden dimensionality of the MLP.

Two key observations are found in the ST block: (1) the majority of the parameter overhead arises from the matrices $\mathbf{U}_{\mathrm{q}}$, $\mathbf{U}_{\mathrm{k}}$, $\mathbf{U}_{\mathrm{v}}$ and $\mathbf{M}_{0}$ in the SSA module, as well as $\mathbf{M}_1$ and $\mathbf{M}_2$ in the MLP module; and (2) as a task-independent, unified backbone for spatio-temporal feature extraction, the ST block presents a promising opportunity to develop universal lightweight strategies applicable across diverse downstream tasks. To this end, we propose two complementary pruning strategies—one unstructured and one structured—aimed at reducing the computational and memory complexity of the standard ST block.

\subsubsection{$\mathbf{L_{1}}$ Norm-based Parameter Sparsification}
Unstructured pruning offers a straightforward and flexible approach to model compression, with high implementation adaptability and demonstrated effectiveness in previous research. In our method, referred to as $\mathbf{L_{1}P}$, we prune by zeroing out a specified proportion of elements with the smallest $\mathrm{L_1}$-norm values within each target weight matrix. This strategy effectively reduces the number of model parameters while preserving the overall functional integrity of the network.

To mitigate potential degradation in model performance, the proposed $\mathbf{L_{1}P}$ method performs pruning independently on each weight matrix. Let $\mathbf{W} \in \mathbb{R}^{m \times n}$ denote a weight matrix, represented as $\mathbf{W} = [w_{ij}]$ for $i = 1, 2, \ldots, m$ and $j = 1, 2, \ldots, n$. The corresponding sparse matrix $\mathbf{\widetilde{W}}$ is constructed through the following steps:
\begin{enumerate}
	\item \textbf{Magnitude Computation:} Compute the element-wise absolute value matrix $\mathbf{A}$, where each element is given by:
	\begin{equation}
	\mathbf{A}_{ij} = \left\| w_{ij} \right\|_1,
	\end{equation}
	for $i = 1, 2, \ldots, m$ and $j = 1, 2, \ldots, n$.
	\item \textbf{Threshold Selection:} Determine the pruning threshold $P_{\mathrm{th}}$ based on a predefined pruning sparsity $p \in [0,1]$ as follows: (a) Sort all elements in $\mathbf{A}$ in ascending order to obtain the sorted sequence $\mathbb{V} = \{v_k \mid k = 1, 2, \ldots, mn\}$; (b) Compute the index $K = \lceil p \cdot mn \rceil$, where $\lceil \cdot \rceil$ denotes the ceiling function, and set the threshold as $P_{\mathrm{th}} = v_{K}$.
	\item \textbf{Pruning Operation:} Apply thresholding to obtain the sparse matrix $\mathbf{\widetilde{W}}$, where each element is updated as:
	\begin{equation}
	\widetilde{w}_{ij} =
	\begin{cases}
	w_{ij}, & \text{if } \left\| w_{ij} \right\|_1 \geq P_{\mathrm{th}}, \\
	0, & \text{otherwise}.
	\end{cases}
	\end{equation}
\end{enumerate}

This unstructured pruning approach retains weights with relatively larger magnitudes, thereby inducing sparsity while maintaining the model’s representational capacity.

\begin{figure}[!t]
	\centering
	\includegraphics[width=1.0\columnwidth]{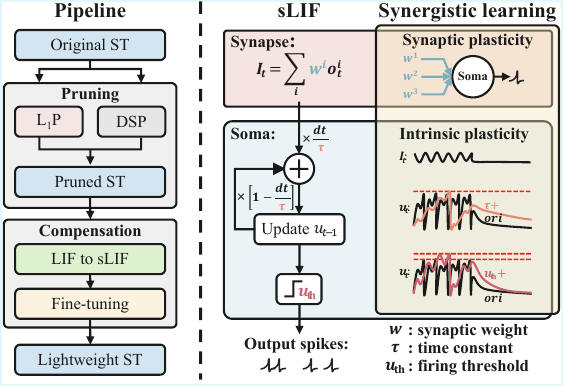}
	\caption{Illustration of the overall pipeline of the proposed method, including the sLIF neuron model and the synergistic learning mechanism.}
	\label{Fig3}
\end{figure}

\subsubsection{Dimension Significance-based Pruning}
Structured pruning represents a more principled and architecture-aware strategy for model compression, wherein entire groups of weights are removed in a structured manner. This approach enables tangible improvements in inference efficiency and model compactness. Within a ST block, the weight matrices function as dimensional projectors, transforming input patch embeddings into a new representational space. Nevertheless, it remains unclear whether the extended dimensionality of patch embeddings is essential, and which specific dimensions can be pruned without significantly degrading performance.

Motivated by the preceding observations, we propose a Dimension Significance-based Pruning~(\textbf{DSP}) method. The \textbf{DSP} approach employs a novel Dimension Value Assessment~(DVA) metric to decompose the original weight matrix into low-rank representations by retaining dimension projectors associated with higher significance scores. Formally, for a weight matrix $\mathbf{W} \in \mathbb{R}^{m \times n}$, which maps an input dimension of size $m$ to an output dimension of size $n$, the DVA metric computes the significance scores $\mathbf{s} \in \mathbb{R}^{n}$ for each output dimension as follows:
\begin{equation}
s_j = \mathrm{DVA}(\mathbf{W}) = \sum_{i=1}^{m} \left\|w_{ij}\right\|_{1}, \quad j = 1, \dots, n.
\end{equation}
As illustrated in Fig.~\ref{Fig2}, for the SSA module, the significance scores $\mathbf{s}_{\mathrm{ssa}} \in \mathbb{R}^{n}$ are computed as the average DVA scores of the weight matrices $\mathbf{U}_{\mathrm{q}}, \mathbf{U}_{\mathrm{k}}, \mathbf{U}_{\mathrm{v}} \in \mathbb{R}^{d \times d}$:
\begin{equation}
\mathbf{s}_{\mathrm{ssa}} = \frac{1}{3} \left[ \mathrm{DVA}(\mathbf{U}_{\mathrm{q}}) + \mathrm{DVA}(\mathbf{U}_{\mathrm{k}}) + \mathrm{DVA}(\mathbf{U}_{\mathrm{v}}) \right].
\end{equation}
Given a predefined pruning sparsity $p \in [0,1]$, the $\lceil p \cdot n \rceil$ dimensions with the lowest significance scores are pruned from the weight matrices $\mathbf{U}_{\mathrm{q}}$, $\mathbf{U}_{\mathrm{k}}$, and $\mathbf{U}_{\mathrm{v}}$, yielding low-rank projectors $\mathbf{\hat{U}}_{\mathrm{q}}, \mathbf{\hat{U}}_{\mathrm{k}}, \mathbf{\hat{U}}_{\mathrm{v}} \in \mathbb{R}^{d \times d_{\downarrow}}$, where $d_{\downarrow} = n - \lceil p \cdot n \rceil$. Subsequently, by pruning the corresponding input dimensions, a low-rank matrix $\mathbf{\hat{M}}_{0} \in \mathbb{R}^{d_{\downarrow} \times d}$ is obtained. Similarly, for the weight matrices $\mathbf{M}_{1}, \mathbf{M}_{2} \in \mathbb{R}^{d \times d}$ in the MLP module, an analogous pruning process is applied to obtain low-rank matrices $\mathbf{\hat{M}}_{1} \in \mathbb{R}^{d \times d_{m\downarrow}}$ and $\mathbf{\hat{M}}_{2} \in \mathbb{R}^{d_{m\downarrow} \times d}$.

\subsection{Synergistic Learning-based Compensation}
The overall pipeline of our proposed method is illustrated in Fig.~\ref{Fig3}, encompassing both synapse pruning and information compensation. Initially, a pre-trained original ST is pruned using either the $\mathrm{L_{1}P}$ or DSP strategy, yielding a pruned model. Subsequently, the original neurons are replaced with sLIF neurons, followed by fine-tuning via synergistic learning to mitigate information loss and produce a lightweight ST block.

Various spiking neuron models have been proposed to mimic the spatio-temporal dynamics of biological neurons. Among them, the LIF model is widely used in SNNs due to its balance between biological fidelity and computational efficiency. With surrogate gradient methods like spatio-temporal backpropagation (STBP), SNNs have shown notable performance in learning complex patterns. However, most studies focus on synaptic plasticity, neglecting the role of intrinsic neuronal parameters in modulating excitability. As illustrated in Fig.~\ref{Fig3}, the membrane time constant determines the rate at which historical information is forgotten, while the firing threshold directly influences the neuronal firing rate. The synergistic learning between synaptic and intrinsic parameters can enhance neuronal heterogeneity, thereby improving the representational capacity of the SNN models.

To bridge this gap, we propose the sLIF model, which integrates synaptic and intrinsic plasticity (IP) into a unified learning framework. The dynamics of the proposed sLIF neuron are governed by the following differential equation:
\begin{equation}
\tau \frac{du_{t}}{dt} = -(u_{t} - u_{\mathrm{rest}}) + I_{t},
\label{eq1}
\end{equation}
where $u_{t}$ denotes the membrane potential at time $t$, $u_{\mathrm{rest}}$ is the resting potential, and $I_{t}$ is the total synaptic input current received by the neuron. In practice, we discretize Eq.~(\ref{eq1}) using the Euler method with a unit time step for computational implementation:
\begin{equation}
u_{t} = u_{t-1} + \frac{-(u_{t-1} - u_{\mathrm{rest}}) + I_{t}}{\tau}.
\label{eq_discrete}
\end{equation}
The generation of output spikes follows a thresholding mechanism:
\begin{equation}
o_{t} = \mathrm{H}(u_{t} - u_{\mathrm{th}}),
\label{eq2}
\end{equation}
where $\mathrm{H}(\cdot)$ denotes the Heaviside step function, and $u_{\mathrm{th}}$ is the firing threshold. Following a spike, the membrane potential is reset using either a soft-reset mechanism:
\begin{equation}
u_{t} = u_{t} - o_{t} u_{\mathrm{th}},
\label{eq_reset1}
\end{equation}
or a hard-reset mechanism:
\begin{equation}
u_{t} = u_{t}(1-o_{t}).
\label{eq_reset2}
\end{equation}
The synaptic input $I_{t}$ is computed as the weighted summation of incoming spikes:
\begin{equation}
I_{t} = \sum_{i} w^{i} o^{i}_{t},
\label{eq3}
\end{equation}
where $w^{i}$ and $o^{i,t}$ represent the synaptic weight and the presynaptic spike from the $i$-th neuron at time $t$, respectively. 

Distinct from prior models, the sLIF neuron treats intrinsic parameters, the membrane time constant $\tau$ and the firing threshold $u_{\mathrm{th}}$, as learnable parameters, and optimizes them alongside the synaptic weights $w$. According to the chain rule, the derivatives of the loss function $Loss$ with respect to $o^{i}_{t,n}$ and $u^{i}_{t,n}$  can be mathematically described as follows:
\begin{equation}
\frac{\partial Loss}{\partial o^{i}_{t,n}}= \frac{\partial Loss}{\partial o^{i}_{t+1,n}}\frac{\partial o^{i}_{t+1,n}}{\partial o^{i}_{t,n}}+\sum_{j=1}^{l(n+1)}\frac{\partial Loss}{\partial o^{j}_{t,n+1}}\frac{\partial o^{j}_{t,n+1}}{\partial o^{i}_{t,n}},  
\label{add1}
\end{equation}
\begin{equation}
\frac{\partial Loss}{\partial u^{i}_{t,n}}= \frac{\partial Loss}{\partial o^{i}_{t,n}}\frac{\partial o^{i}_{t,n}}{\partial u^{i}_{t,n}}+\frac{\partial Loss}{\partial o^{i}_{t+1,n}}\frac{\partial o^{i}_{t+1,n}}{\partial u^{i}_{t,n}}.  
\label{add2}
\end{equation}
Here, $o^{i}_{t,n}$ and $u^{i}_{t,n}$ represent the output spike and membrane potential of the $i$-th neuron in the $n$-th layer at time $t$. $l(n+1)$ denotes the number of neurons in the $(n+1)$-th layer. Based on equations~(\ref{add1}) and (\ref{add2}), we finally obtain the derivatives with respect to the synaptic weight, firing threshold and membrane time constant:   
\begin{equation}
\frac{\partial Loss}{\partial w^{i}_{n}}= \sum_{t=1}^{T} \frac{\partial Loss}{\partial u^{i}_{t, n}} \frac{\partial u^{i}_{t, n}}{\partial w^{i}_{n}}, 
\label{add3}
\end{equation}
\begin{equation}
\frac{\partial Loss}{\partial u_{\text{th},n}}= \sum_{t=1}^{T}  \frac{\partial Loss}{\partial o^{i}_{t, n}} \frac{\partial o^{i}_{t, n}}{\partial u_{\text{th},n}}, 
\label{add4}
\end{equation}
\begin{equation}
\frac{\partial Loss}{\partial \tau_{n}}= \sum_{t=1}^{T}  \frac{\partial Loss}{\partial u^{i}_{t, n}} \frac{\partial u^{i}_{t, n}}{\partial \tau_{n}}. 
\label{add5}
\end{equation}

To mitigate the performance degradation induced by synapse pruning, the LIF neurons in the ST blocks are replaced with the proposed sLIF neuron model. Subsequently, the pruned model is fine-tuned using synergistic learning over a small number of epochs. To facilitate efficient adaptation, the synaptic and intrinsic parameters are initialized by transferring them from the original model.

\begin{figure}[!t]
	\centering
	\includegraphics[width=1.0\columnwidth]{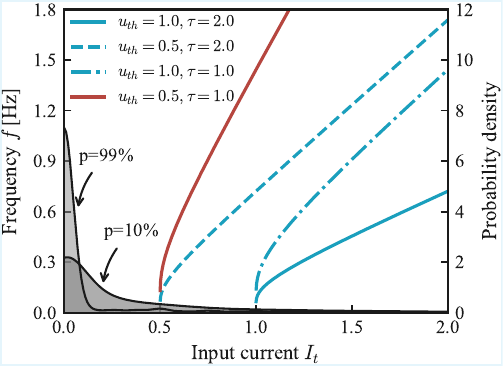}
	\caption{Response re-alignment via synergistic plasticity of $u_{th}$ and $\tau$. Decreasing $u_{th}$ performs a horizontal shift of the f-I curve to the left, enabling the neuron to respond to weak inputs induced by pruning. Varying $\tau$ modulates the curvature and sensitivity of the response. The joint optimization of both parameters allows the sLIF neuron to maintain robust firing rates and information flow even under extreme pruning rates.}
	\label{Fig_add1}
\end{figure}

\subsection{Theoretical Analysis}
\subsubsection{Synergistic Gradient Restoration}
From an optimization perspective, synapse pruning induces a distribution mismatch between the membrane potential $u_t$ and the sensitive region of the surrogate gradient $\sigma'(u_t - u_{\text{th}})$. The synergistic learning of both the firing threshold $u_{\text{th}}$ and the membrane time constant $\tau$ is necessary to effectively restore the gradient flow adaptively.

Let $\mathcal{S} = \{u \mid \sigma'(u - u_{\text{th}}) > \epsilon\}$ denote the gradient-sensitive window centered at $u_{\text{th}}$. Due to the reduced input current $\mathbb{E}[I_t]$ after pruning, the membrane potential $u_t$ drifts outside $\mathcal{S}$ (i.e., $u_t \ll u_{\text{th}}$), causing gradient vanishing. The restoration of gradients requires realigning the potential distribution with $\mathcal{S}$, which necessitates two complementary mechanisms. The synergistic learning of $\{u_{\text{th}}, \tau\}$ functions as a dynamic affine transformation: $\tau$ amplifies the weak signals to a robust range (scaling), while $u_{\text{th}}$ fine-tunes the decision boundary (shifting). This coupling ensures that the membrane potential $u_t$ is optimally re-mapped into the gradient-sensitive window $\mathcal{S}$:
\begin{equation}
\min_{u_{\text{th}}, \tau} \sum_{t} \left| u_t(\tau) - u_{\text{th}} \right| \implies \sigma'(u_t - u_{\text{th}}) \gg 0.
\end{equation}
Thus, the joint optimization guarantees robust gradient recovery and efficient fine-tuning for pruned SNNs.

\subsubsection{ Response Re-alignment via Synergistic Plasticity}
From the perspective of neural dynamics, synapse pruning creates an ``Input Distribution Shift" that impairs the information transmission capability of spiking neurons. We analyze this phenomenon using the frequency-current (f-I) curve and demonstrate how the synergistic plasticity of sLIF restores neuronal responsiveness. Based on the sLIF dynamics described in Eq. (11), the steady-state firing rate $f(I)$ for a constant input current $I$ can be derived as:
\begin{equation}
f(I) = \left[ t_{\text{ref}} + \tau \ln \left( \frac{I}{I - (u_{\text{th}} - u_{\text{rest}})} \right) \right]^{-1},
\end{equation}
where $t_{\text{ref}}$ is the refractory period. The term $I_{\text{rheo}} = u_{\text{th}} - u_{\text{rest}}$ represents the rheobase, defined as the minimum current required to trigger a spike. As shown in Fig.~\ref{Fig_add1}, synapse pruning significantly reduces the magnitude of the integrated current $I_t$, shifting its probability distribution towards zero. Consequently, the majority of inputs fall below the rheobase of a standard LIF neuron with a fixed threshold, resulting in a silent regime where $f(I) \to 0$ and the forward information flow is blocked. 

The sLIF neuron recovers excitability by reshaping the f-I curve through the synergistic adaptation of $u_{\text{th}}$ and $\tau$. Specifically, learning a lower $u_{\text{th}}$ decreases $I_{\text{rheo}}$, which geometrically shifts the f-I curve to the left, ensuring that even weak pruned currents can exceed the firing threshold. Simultaneously, the membrane time constant $\tau$ acts as a scaling factor that modulates the curvature and gain of the response. While threshold adaptation ensures the neuron can fire, time constant adaptation regulates how it encodes signals. This synergistic coupling allows the sLIF neuron to dynamically match its dynamic range with the pruned input distribution, preserving robust information transmission in lightweight models.

\begin{table}[!t]
	\caption{Configuration Parameters for Experiments}
	\centering
	\setlength{\tabcolsep}{1.5mm}{
		\begin{tabular}{rcccc}
			\hline
			Hyper-parameter &    ImageNet    &   CIFAR10    &  CIFAR10-DVS   &     ADE20K     \\ \hline
			Epochs/iterations &       50       &      50      &       56       &     200000     \\
			Warmup epochs &       20       &      0       &       10       &      1500      \\
			Batch size &       64       &     128      &       16       &       16       \\
			Optimizer &     AdamW      &    AdamW     &     AdamW      &     AdamW      \\
			Initial learning rate &     0.0001     &   0.00001    &     0.0001     &     0.001      \\
			Learning rate decay &     Cosine     &    Cosine    &     Cosine     &    LinearLR    \\
			Weight decay &      0.05      &     0.06     &      0.06      &     0.005      \\
			Time steps &       4        &      4       &       16       &       4        \\
			Resolution & 224$\times$224 & 32$\times$32 & 128$\times$128 & 512$\times$512 \\
			Patch size &       16       &      4       &       16       &       16       \\ \hline
	\end{tabular}}
	\label{Tab1}
\end{table}

\begin{table*}[!t]
	\caption{Classification results on the ImageNet-100 dataset. The symbol $^{\dag}$ denotes the baseline model. The terms p and CR indicate the pruning sparsity and compression ratio, respectively. The notation ‘/’ separates values measured on the entire model and on the ST blocks.}
	\centering
	\begin{tabular}{cccccc}
		\hline
		                      Method                       &     Architecture      & p [\%] &   CR [\%]   &     Param [M]      & Accuracy [\%]  \\ \hline
		       LOCALZO+TET~\cite{mukhoty2023direct}        &      SEWResNet34      &   -    &      -      &       63.47        &     78.58      \\
		           IM-SNN~\cite{hasssan2024snn}            &       Resnet34        &   -    &      -      &       21.27        &     74.42      \\
		        IMP+TET~\cite{shen2024rethinking}          &     SEW-ResNet18      &   -    &      -      &       63.47        &     78.70      \\
		EfficientLIF-Net~\cite{10.3389/fnins.2023.1230002} &         VGG16         &   -    &      -      &       23.52        &     73.22      \\ \hline
		  Spikformer$^{\dag}$~\cite{zhou2023spikformer}    & Spikformer-8-512-2048 &   -    &      -      &    29.24/25.17     & \textbf{79.36} \\ \hline
		          $\mathrm{L_{1}P}$+sLIF~(ours)            & Spikformer-8-512-2048 &   90   & 77.43/89.99 &     6.60/2.52      &  76.22(-3.14)  \\
		          $\mathrm{L_{1}P}$+sLIF~(ours)            & Spikformer-8-512-2048 &   99   & 85.19/99.00 & \textbf{4.33/0.25} & 62.76(-16.60 ) \\
		                 DSP+sLIF~(ours)                   &  Spikformer-8-48-204  &   90   & 77.63/90.46 &     6.54/2.40      &  76.88(-2.48)  \\
		                 DSP+sLIF~(ours)                   &   Spikformer-8-8-20   &   99   & 85.05/99.05 &     4.37/0.24      & 62.76(-16.60)  \\ \hline
	\end{tabular}
	\label{Tab2}
\end{table*}

\begin{table*}[!t]
\caption{Classification results on the CIFAR10 and CIFAR10-DVS datasets. The symbol $^{\dag}$ denotes the baseline model. The terms p and CR indicate the pruning sparsity and compression ratio, respectively. The notation ‘/’ separates values measured on the entire model and on the ST blocks.}
\centering
\begin{tabular}{ccccccc}
	\hline
                 Dataset                    &                     Method                     &     Architecture      & p [\%] &   CR [\%]   &     Param [M]      & Accuracy [\%]  \\ \hline
 \multirow{15}{*}{\rotatebox{90}{CIFAR10}}   &        STBP-tdBN~\cite{zheng2021going}         &       ResNet-19       &   -    &      -      &       11.17        &     92.92      \\
                                            &            STP~\cite{ma2025spiking}            &       ResNet18        &   -    &      -      &       63.47        &     94.86      \\
                                            &         STL-SNN~\cite{sun2023synapse}          &        ConvFC         &   -    &      -      &       11.37        &     92.42      \\ \cline{2-7}
                                            & Spikformer$^{\dag}$~\cite{zhou2023spikformer}  & Spikformer-4-384-1536 &   -    &      -      &     9.32/7.08      &     94.76      \\
                                            &       Spikingformer~\cite{spikingformer}       & Spikformer-4-384-1536 &   -    &      -      &     9.32/7.08      &     95.81      \\
                                            &          CML~\cite{zhou2023enhancing}          & Spikformer-4-384-1536 &   -    &      -      &     9.32/7.08      & \textbf{96.04} \\ \cline{2-7}
                                            & \multirow{5}{*}{$\mathrm{L_{1}P}$+sLIF~(ours)} & Spikformer-4-384-1536 &   20   & 18.03/23.73 &     7.64/5.40      &  94.72(-0.04)  \\
                                            &                                                & Spikformer-4-384-1536 &   40   & 32.51/42.80 &     6.29/4.05      &  94.78(+0.02)  \\
                                            &                                                & Spikformer-4-384-1536 &   60   & 47.00/61.86 &     4.94/2.70      &  94.61(-0.15)  \\
                                            &                                                & Spikformer-4-384-1536 &   80   & 60.62/79.94 &     3.67/1.42      &  93.94(-0.82)  \\
                                            &                                                & Spikformer-4-384-1536 &   90   & 68.24/89.97 & \textbf{2.96/0.71} &  92.32(-2.44)  \\ \cline{2-7}
                                            &        \multirow{5}{*}{DSP+sLIF~(ours)}        & Spikformer-4-312-1228 &   20   & 17.60/23.31 &     7.68/5.43      &  94.60(-0.16)  \\
                                            &                                                & Spikformer-4-240-921  &   40   & 31.76/41.95 &     6.36/4.11      &  94.36(-0.40)  \\
                                            &                                                & Spikformer-4-156-614  &   60   & 46.67/61.58 &     4.97/2.72      &  94.03(-0.73)  \\
                                            &                                                &  Spikformer-4-84-307  &   80   & 60.19/79.38 &     3.71/1.46      &  93.14(-1.62)  \\
                                            &                                                &  Spikformer-4-48-153  &   90   & 67.60/89.12 &     3.02/0.77      &  92.23(-2.53)  \\ \hline
	\multirow{15}{*}{\rotatebox{90}{CIFAR10-DVS}} &          TET~\cite{deng2022temporal}           &        VGGSNN         &   -    &      -      &        9.54        &     77.33      \\
                                            &            STP~\cite{ma2025spiking}            &         VGG11         &   -    &      -      &       113.00       &     78.50      \\
                                            &         STL-SNN~\cite{sun2023synapse}          &        ConvFC         &   -    &      -      &        1.53        &     77.30      \\ \cline{2-7}
                                            & Spikformer$^{\dag}$~\cite{zhou2023spikformer}  & Spikformer-2-256-1024 &   -    &      -      &     2.59/1.58      &     79.40      \\
                                            &       Spikingformer~\cite{spikingformer}       & Spikformer-2-256-1024 &   -    &      -      &     2.59/1.58      & \textbf{81.30} \\
                                            &          CML~\cite{zhou2023enhancing}          & Spikformer-2-256-1024 &   -    &      -      &     2.57/1.58      &     79.20      \\ \cline{2-7}
                                            & \multirow{5}{*}{$\mathrm{L_{1}P}$+sLIF~(ours)} & Spikformer-2-256-1024 &   20   & 14.67/24.05 &     2.21/1.20      &  80.00(+0.60)  \\
                                            &                                                & Spikformer-2-256-1024 &   40   & 26.25/43.04 &     1.91/0.90      &  80.10(+0.70)  \\
                                            &                                                & Spikformer-2-256-1024 &   60   & 37.84/62.03 &     1.61/0.60      &  80.40(+1.00)  \\
                                            &                                                & Spikformer-2-256-1024 &   80   & 48.65/79.75 &     1.33/0.32      &  78.00(-1.40)  \\
                                            &                                                & Spikformer-2-256-1024 &   90   & 54.83/25.95 &     1.17/0.16      &  76.30(-3.10)  \\ \cline{2-7}
                                            &        \multirow{5}{*}{DSP+sLIF~(ours)}        & Spikformer-2-192-819  &   20   & 15.83/43.67 &     2.18/1.17      &  78.90(-0.50)  \\
                                            &                                                & Spikformer-2-144-614  &   40   & 26.64/62.66 &     1.90/0.89      &  78.90(-0.50)  \\
                                            &                                                &  Spikformer-2-96-409  &   60   & 38.22/80.38 &     1.60/0.59      &  78.90(-0.50)  \\
                                            &                                                &  Spikformer-2-48-204  &   80   & 49.03/80.38 &     1.32/0.31      &  77.20(-2.20)  \\
                                            &                                                &  Spikformer-2-16-102  &   90   & 55.60/91.14 & \textbf{1.15/0.14} &  77.30(-2.10)  \\ \hline
\end{tabular}
\label{Tab3}
\end{table*}

\section{Experiments and Results}\label{Experiments and Results}
\subsection{Experimental Settings}
To evaluate the effectiveness of our method, experiments are conducted on both static and neuromorphic image datasets, ImageNet~\cite{5206848}, CIFAR-10~\cite{krizhevsky2009learning}, and CIFAR10-DVS~\cite{li2017cifar10}. To further validate the generalization capability in downstream tasks, semantic segmentation is performed on the ADE20K dataset~\cite{zhou2017scene}. For a fair comparison, our method are directly applied to state-of-the-art~(SOTA) pre-trained models implemented using the PyTorch and SpikingJelly frameworks~\cite{zhou2023spikformer, yao2024spikedriven, doi:10.1126/sciadv.adi1480}. All experiments are executed on 4 NVIDIA A800 GPUs. 

\subsubsection{Datasets}
To comprehensively evaluate the performance and generalization capability of the proposed method across both static and neuromorphic vision tasks, we utilize a diverse set of benchmark datasets. Below, we provide detailed descriptions of each dataset, highlighting their composition, key characteristics, and relevance to our experimental settings. 

ImageNet~\cite{5206848} is a large-scale static image dataset widely used for object recognition tasks, comprising approximately 1.28 million training images and 50,000 validation images across 1,000 classes. In our experiments, all images are resized to a resolution of 224$\times$224 pixels for consistency. To evaluate the classification performance of our method on high-resolution static images, we adopt the ImageNet-100 subset~\cite{imagenet100pytorch}, where the classes are selected following prior work. This subset serves as a benchmark for comparison with state-of-the-art (SOTA) models in conventional computer vision tasks. Additionally, the full ImageNet-1K dataset is employed to assess the scalability of our method under the scaling-law setting. 

CIFAR-10~\cite{krizhevsky2009learning} is a static image dataset consisting of 50,000 training images and 10,000 test images, each with a resolution of 32$\times$32 pixels and evenly distributed across 10 classes. It is used to evaluate the effectiveness of our method on low-resolution static images, offering a lightweight yet challenging benchmark for classification tasks. 

CIFAR10-DVS~\cite{li2017cifar10} is a neuromorphic dataset derived from the original CIFAR-10, recorded using a dynamic vision sensor (DVS). It comprises event streams from 10,000 samples spanning the same 10 classes as CIFAR-10, where each sample is represented as a sequence of address-event representations (AER) instead of conventional image frames. With temporal resolution on the order of microseconds, this dataset is well-suited for evaluating the performance of our method on event-based, neuromorphic vision tasks. In our experiments, each event stream is temporally averaged and segmented into 16 discrete time steps. 

ADE20K~\cite{zhou2017scene} is a widely used dataset for semantic segmentation, comprising 20,210 training images and 2,000 validation images annotated with 150 semantic categories. For segmentation tasks, images are typically resized to a resolution of 512$\times$512 pixels. Owing to its diverse scene compositions and fine-grained annotations, ADE20K serves as a benchmark to assess the generalization capability of our method on downstream tasks, particularly semantic segmentation.

\subsubsection{Configuration Details}
Tab.~\ref{Tab1} summarizes the key configuration parameters used for each dataset in our experiments. The source code will be released upon acceptance at https://github.com/GuoLab-UESTC/EfficientST.
\begin{figure}[!t]
	\centering
	\includegraphics[width=1.0\columnwidth]{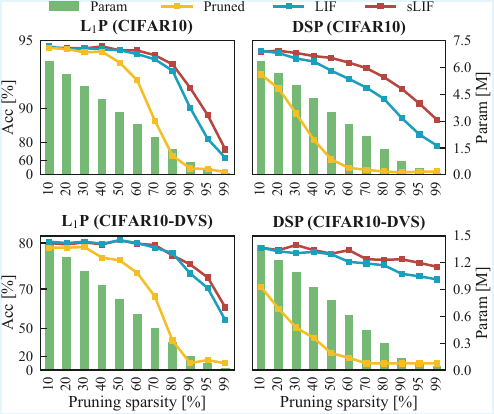}
	\caption{Performance comparison of the proposed lightweight models (sLIF) and LIF-compensated models (LIF) under varying pruning sparsity on the CIFAR10 and CIFAR10-DVS datasets. The parameter counts (Param) and pruned accuracies (Pruned) of the baseline models are also provided to demonstrate the effectiveness of the proposed method.}
	\label{Fig4}
\end{figure}

\subsection{Comparisons with SOTA Methods}
\subsubsection{Static Datasets Classification}
We compare our lightweight models with other SNNs on the ImageNet-100 dataset (Tab.~\ref{Tab2}). The baseline Spikformer-8-512-2048 model consists of eight ST blocks with $d=512$ and $d_{\mathrm{m}}=2048$. After pruning with a sparsity level of $p=90\%$, the parameter count is reduced to below $7$M, achieving a compression ratio (CR) exceeding $77\%$, while incurring only a minor accuracy drop ($-3.14\%$ for $\mathrm{L_{1}P}$+sLIF and $-2.48\%$ for DSP+sLIF). Considering only the ST blocks, the CR exceeds $90\%$, with parameters reduced to below $2.5$M. When sparsity is further increased to $p=99\%$, the accuracy decreases by about $16\%$. Nevertheless, given that the parameter count of the ST blocks is reduced from $24$M to only $0.25$M (for $\mathrm{L_{1}P}$+sLIF) or $0.24$M (for DSP+sLIF), this trade-off remains acceptable for lightweight deployment. Moreover, compared to ResNet- or VGG-based convolutional models with comparable accuracy, our lightweight models demonstrate substantially higher parameter efficiency.

On the CIFAR10 dataset~(Tab.~\ref{Tab3}), ST-based models such as Spikformer and CML exhibit superior performance, with CML attaining the highest accuracy of $96.04\%$ when employing the Spikformer-4-384-1536 architecture. Utilizing our proposed $\mathrm{L_{1}P}$+sLIF method, an accuracy of $93.94\%$ is achieved under a sparsity level of $p=80\%$, accompanied by a substantial compression ratio of $60.62\%$ for the entire model and $79.94\%$ for the ST modules, and a significantly reduced parameter count of $3.67$M and $1.42$M, respectively. When the sparsity is further increased to $p=90\%$, the model maintains a competitive accuracy of $92.32\%$, while the parameter count is further reduced to $2.96$M (entire model) and $0.71$M (ST modules). Similarly, the proposed DSP+sLIF strategy also demonstrates strong performance with enhanced parameter efficiency, achieving $93.14\%$ ($92.23\%$) accuracy with only $3.71$M ($3.02$M) parameters under $p=80\%$ ($p=90\%$).

\subsubsection{Neuromorphic Datasets Classification}
On the CIFAR10-DVS dataset, the proposed lightweight models also demonstrate competitive performance, as summarized in Tab.~\ref{Tab3}. The baseline Spikformer model achieves an accuracy of $80.90\%$ with $2.59$M parameters ($1.58$M in ST modules). By applying the $\mathrm{L_{1}P}$+sLIF method under a sparsity level of $p=80\%$, the model is compressed to $1.33$M parameters  ($0.32$M in ST modules) while maintaining a commendable accuracy of $78.00\%$. Similarly, the DSP+sLIF strategy yields a compact Spikformer-2-48-204 model with $1.32$M parameters ($0.31$M in ST modules) and an accuracy of $77.20\%$ at the same sparsity level. Performance remains comparable when sparsity is further increased.

\begin{figure}[!t]
	\centering
	\includegraphics[width=1.0\columnwidth]{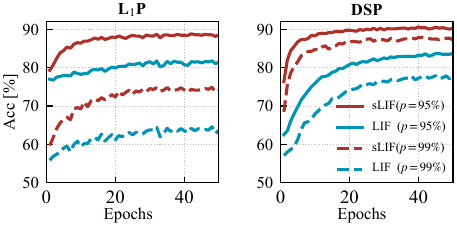}
	\caption{Comparison of convergence curves of pruned models with LIF- and sLIF-based compensation on the CIFAR-10 dataset during fine-tuning phase.}
	\label{Fig5}
\end{figure}

\begin{figure}[!t]
	\centering
	\includegraphics[width=1.0\columnwidth]{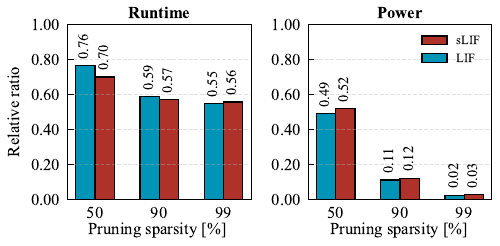}
	\caption{Relative ratio of runtime and theoretical power consumption of the lightweight models compared to the baseline models.}
	\label{Fig6}
\end{figure}

\begin{figure}[!t]
	\centering
	\includegraphics[width=1.0\columnwidth]{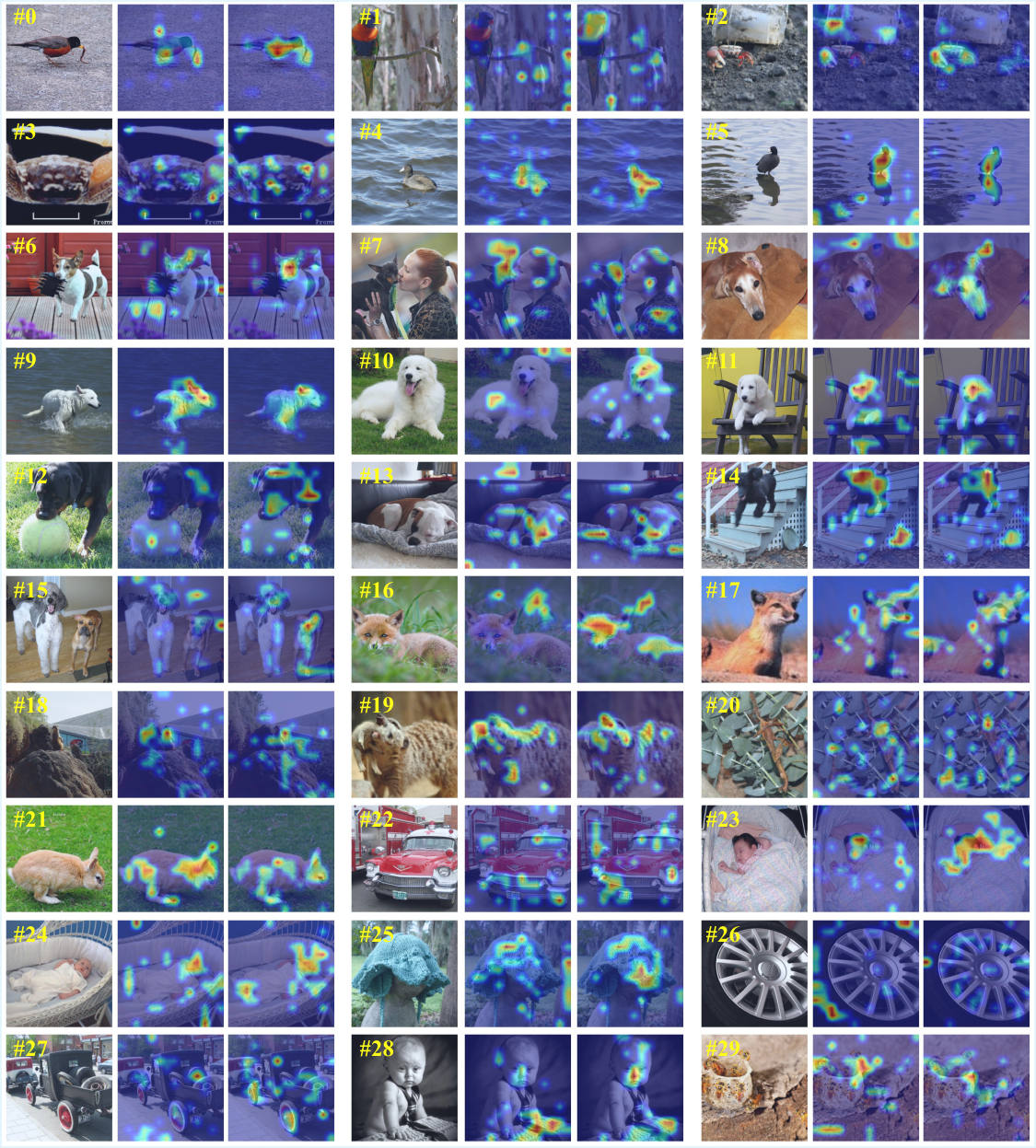}
	\caption{Representative examples of attention maps generated by the original~(middle) and lightweight~(right) models on the ImageNet-100 dataset. Higher attention values are indicated in deep red.}
	\label{Fig7}
\end{figure}

\begin{algorithm}[!t]
	\caption{Attention Rollout Visualization}
	\label{attention-rollout}
	\begin{algorithmic}[1]
		\REQUIRE Attention maps $\{\mathbf{A}_l\}_{l=1}^L$, where $\mathbf{A}_l \in \mathbb{R}^{H \times P \times P}$, $H$ is the number of heads, $P$ is the number of patches
		\ENSURE Visualization mask $\mathbf{M} \in \mathbb{R}^{P}$
		\STATE Initialize $\mathbf{R} = \mathbf{I}_P$ \quad \text{(identity matrix of size $P$)}
		\FOR{$l = 1$ to $L$}
		\STATE Compute fused attention: $\mathbf{A}_l^{\text{fused}} = \text{mean}\big(\mathbf{A}_l, \text{axis=0}\big) \in \mathbb{R}^{P \times P}$ 
		\STATE Add residual: $\mathbf{A}_l' = \mathbf{A}_l^{\text{fused}} + \mathbf{I}_P$
		\STATE Normalize rows: $\mathbf{A}_l' = \mathbf{A}_l' \oslash \big(\mathbf{A}_l'.sum(\text{axis=1}, \text{keepdims}=\texttt{True}\big)$
		\STATE Update rollout: $\mathbf{R} = \mathbf{A}_l' \cdot \mathbf{R}$
		\ENDFOR
		\STATE Compute mask: $\mathbf{M} = \text{mean}\big(\mathbf{R}, \text{axis=0}\big) \in \mathbb{R}^P$
		\STATE Apply threshold: Set the smallest \texttt{discard\_ratio} proportion of values in $\mathbf{M}$ to zero
		\STATE Reshape and upsample $\mathbf{M}$ to the original image size for visualization
	\end{algorithmic}
\end{algorithm}

\subsection{Ablation Analysis}
\subsubsection{Performance Under Different Levels of Sparsity}
To comprehensively evaluate the performance of the proposed lightweight strategies, we perform pruning experiments under different sparsity levels~(denoted as `sLIF'). For comparison, we also assess the accuracy of the original model after pruning~(denoted as `Pruned'). Additionally, to evaluate the contribution of synergistic learning, we conduct pruning experiments while retaining the original LIF neurons~(denoted as `LIF'). As illustrated in Fig.~\ref{Fig4}, model accuracy generally declines after pruning, with greater sparsity levels resulting in more pronounced performance degradation. Compared to the unstructured pruning strategy~($\mathrm{L_{1}P}$), the structured pruning approach~(DSP) causes more substantial performance deterioration, which is consistent with prior findings. Nonetheless, fine-tuning is consistently an efficient strategy to recover performance. Notably, models fine-tuned with synergistic learning exhibit superior performance recovery, with the benefits of synergistic learning becoming increasingly evident at higher sparsity levels.

In addition to superior compensation performance, the convergence speed during the fine-tuning process is a critical factor in evaluating the effectiveness of a compensation strategy. We present the accuracy curves for the `sLIF' and `LIF' models under pruning sparsity of $p = 99\%$ and $p = 95\%$ on the CIFAR10 dataset, as shown in Fig.~\ref{Fig5}. In our experiments, we configured the fine-tuning process to run for 50 epochs. Across models pruned with various strategies, the sLIF approach consistently enables faster convergence during fine-tuning. Notably, this advantage is particularly pronounced for models pruned using the DSP method. Specifically, models fine-tuned with sLIF achieve convergence in approximately 20 epochs, whereas the ablation group employing LIF neurons requires around 40 epochs to reach convergence.

\subsubsection{Impact on Model Inference Performance}
Inference latency and energy consumption represent two fundamental metrics for assessing the inference performance of models. Since structured pruning facilitates improved hardware accessibility, we evaluate the average batch inference runtime and theoretical power consumption of the DSP+LIF and DSP+sLIF models under pruning sparsity levels of $p = 50\%$, $90\%$, and $99\%$, respectively. The relative ratios of the lightweight models compared to their corresponding baseline models are presented in Fig.~\ref{Fig6}.

As the number of parameters decreases, the inference runtime of the models consistently declines. In particular, the most compact pruned architecture, Spikformer-4-12-15, achieves an inference time that is approximately $50\%$ of that of the baseline model. This acceleration is primarily attributed to the lower computational complexity of the lightweight models, offering promising potential for improving offline performance on edge hardware platforms.

SNN models predominantly utilize sparse accumulate operations as their primary computational units, resulting in significantly lower power consumption compared to traditional artificial neural networks. Additionally, the overall energy consumption of SNNs is influenced by both the model architecture and the neuronal firing rate. In this study, we estimate theoretical energy consumption following the method used in prior work~\cite{sun2024reliable}. As shown in Fig.~\ref{Fig6}, the relative energy consumption of pruned models, compared to the baseline, shows a positive correlation with the number of model parameters. Furthermore, sLIF models demonstrate slightly higher energy consumption than their LIF counterparts.

\begin{table}[!t]
	\caption{Performance on the large-scale ImageNet-1k dataset.}
	\centering
	\setlength{\tabcolsep}{3.4mm}{
		\begin{tabular}{ccccc}
			\hline
			Sparsity  & baseline & $p=30\%$ & $p=50\%$ & $p=90\%$ \\ \hline
			Param [M] &  29.71   &   22.05    &   17.12    &    7.00    \\
			Acc [\%]  &  72.86   &   67.70    &   65.02    &   56.00    \\ \hline
	\end{tabular}}
	\label{Tab4}
\end{table}

\subsubsection{Visualization of Attention Maps}
To gain insight into the decision-making process of our ST-based model, we utilize an adapted version of the Attention Rollout method~\cite{Abnar2020QuantifyingAF}. This technique enables us to aggregate attention across all layers and visualize the regions of the input image that significantly influence the predictions of the model. Specifically, for each layer, we compute the fused attention matrix by averaging across the attention heads (Notably, all attention matrices are selected at the final time step). We then incorporate residual connections by adding the identity matrix and normalize each row to ensure it sums to one. The rollout matrix is obtained by recursively multiplying these processed attention matrices from the first to the last layer. Given that our model employs global average pooling for classification instead of a dedicated class token, we derive the importance of each patch by averaging the rollout matrix across all tokens. To emphasize the most salient regions, we apply a thresholding mechanism that sets the smallest attention values to zero based on a predefined discard ratio (set to $0.85$ in our work). Finally, the resulting mask is upsampled to the original image dimensions and overlaid on the input image to produce an intuitive visualization. The entire pipeline is shown in Algorithm~\ref{attention-rollout}. 

In the DSP+sLIF models, the dimensions of the $Q$, $K$, and $V$ representations are reduced to extremely low levels. For instance, pruning the original `Spikformer-8-512-2048' architecture~(Tab.~\ref{Tab2}) with a pruning sparsity of $p = 90\%$ results in a lightweight `Spikformer-8-48-204' model, where only 48 dimensions are used to represent the patch embeddings. To assess the representational capacity of the lightweight model, we compare the attention maps at the final time step between the baseline (`Spikformer-8-512-2048') and pruned models (`Spikformer-8-48-204') on the ImageNet-100 dataset (shown in Fig.~\ref{Fig7}). Despite the reduced dimensionality, the lightweight model still effectively captures image regions relevant to classification tasks. These results demonstrate that our compensation strategy is an efficient and effective approach for compressing ST-based models. 

\begin{table}[!t]
	\caption{Semantic segmentation performance on the ADE20K dataset.}
	\centering
	\setlength{\tabcolsep}{1.8mm}{
		\begin{tabular}{cccccc}
			\hline
			Method                  &      ST      & p [\%] & Param [M] & mIoU [\%] & mAcc [\%] \\ \hline
			MetaFormer~\cite{yu2022metaformer}    &   $\times$   &   -    &   15.50   &   32.90   &     -     \\
			DeeplabV3~\cite{zhang2022resnest}     &   $\times$   &   -    &   68.10   &   42.10   &     -     \\
			SDTv2$^{\dag}$~\cite{yao2024spikedriven} & $\checkmark$ &   -    &   9.42    &   30.14   &   42.22   \\ \hline
			DSP+sLIF~(ours)              & $\checkmark$ &   50   &   6.52    &   29.71   &   41.65   \\
			DSP+sLIF~(ours)              & $\checkmark$ &   75   &   5.36    &   27.60   &   38.51   \\
			DSP+sLIF~(ours)              & $\checkmark$ &   90   &   4.70    &   26.69   &   37.73   \\ \hline
	\end{tabular}}
	\label{Tab5}
\end{table}

\begin{figure}[t]
	\centering
	\includegraphics[width=1.\columnwidth]{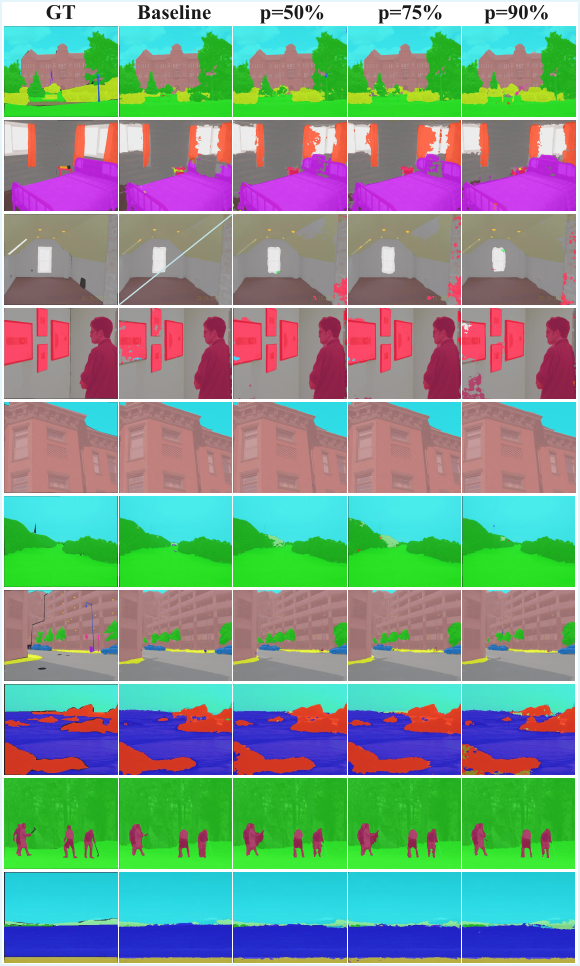}
	\caption{Qualitative semantic segmentation results on the ADE20K dataset.}
	\label{Fig10}
\end{figure}

\subsection{Performance on the Large-Scale Dataset}
We conduct experiments with the DSP+sLIF method on the more challenging ImageNet-1K dataset (Tab.~\ref{Tab4}). The model architecture and experimental settings are aligned with those used for ImageNet-100. Compared to ImageNet-100, a more significant performance drop is observed at high sparsity levels (e.g., $16.86\%$ on ImageNet-1K vs. $2.48\%$ on ImageNet-100 at $p=90\%$). This degradation is theoretically reasonable, as larger-scale datasets require larger model size according to scaling laws~\cite{dosovitskiy2021an}. Nevertheless, the DSP+sLIF method remains effective, achieving competitive performance under extreme compression even on large-scale datasets.

\subsection{Performance on the Downstream Tasks}
To further demonstrate the effectiveness of the proposed DSP+sLIF method on complex downstream tasks, we apply it to semantic segmentation. The ST-based SDTv2 models~\cite{yao2024spikedriven} are used as the baseline, and the DSP+sLIF strategy is employed to compress the original models with pruning sparsity levels of $p=50\%$, $p=75\%$, and $p=90\%$. As shown in Tab.~\ref{Tab5}, the resulting lightweight ST-based semantic segmentation models achieve $26.69\%$ mIoU and $37.73\%$ mAcc performance while requiring only $4.70$M parameters. Moreover, appropriately enlarging the size of the lightweight models further mitigates the performance loss. When the baseline model is compressed with a sparsity level of $p=50\%$, the mIoU~(mAcc) decreases only slightly by $0.43\%$~($0.57\%$).

As illustrated in Fig.~\ref{Fig10}, the semantic segmentation performance of the lightweight model ($p=90\%$) exhibits a slight degradation compared to the original model. Nonetheless, considering that the lightweight model utilizes only half the number of parameters, its performance remains competitive and acceptable for practical applications.

\begin{figure}[htbp]
	\centering
	\includegraphics[width=1.0\columnwidth]{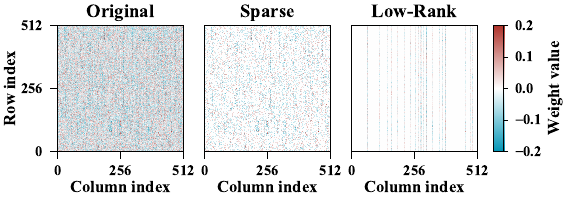}
	\caption{Visualization of the original, sparse weight matrices and low-rank weight matrices. Each pixel represents a value of weight elements and heavy color is correlated with a more higher absolute value.}
	\label{Fig11}
\end{figure}

\begin{table}[t]
	\caption{Comparison of models pruned with proposed methods and random pruning.}
	\centering
	\setlength{\tabcolsep}{3.0mm}{
		\begin{tabular}{c|cc|cc}
			\hline
			\multirow{2}{*}{ p [\%]} & \multicolumn{2}{c|}{\textbf{Sparse weight matrix}} & \multicolumn{2}{c}{\textbf{Low-rank weight matrix}} \\
			& Random [\%] &        $\mathrm{L_{1}P}$ [\%]        & Random [\%] &               DSP [\%]                \\ \hline
			$10$         &    93.97    &                94.72                 &    82.50    &                 93.33                 \\
			$20$         &    90.74    &                94.67                 &    77.63    &                 92.26                 \\
			$30$         &    84.36    &                94.53                 &    69.16    &                 89.16                 \\
			$40$         &    71.22    &                94.56                 &    62.26    &                 81.33                 \\
			$50$         &    52.68    &                94.03                 &    47.00    &                 61.76                 \\
			$60$         &    35.50    &                92.92                 &    39.77    &                 36.81                 \\
			$70$         &    31.29    &                87.72                 &    38.81    &                 28.23                 \\
			$80$         &    29.03    &                67.93                 &    23.15    &                 19.02                 \\
			$90$         &    18.77    &                34.89                 &    17.42    &                 14.17                 \\
			$95$         &    13.00    &                31.74                 &    22.52    &                 18.61                 \\
			$99$         &    10.49    &                16.01                 &    17.83    &                 18.88                 \\ \hline
	\end{tabular}}
	\label{Tab6}
\end{table}

\begin{figure}[!t]
	\centering
	\includegraphics[width=1.\columnwidth]{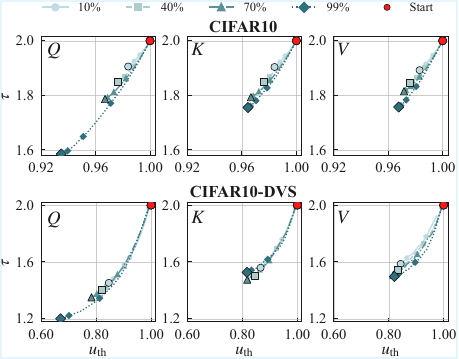}
	\caption{Evolution of IP parameters of the Q, K and V embeddings under different pruning sparsity. `Start' indicates the initial parameter values.}
	\label{Fig8}
\end{figure}

\begin{figure}[!t]
	\centering
	\includegraphics[width=1.\columnwidth]{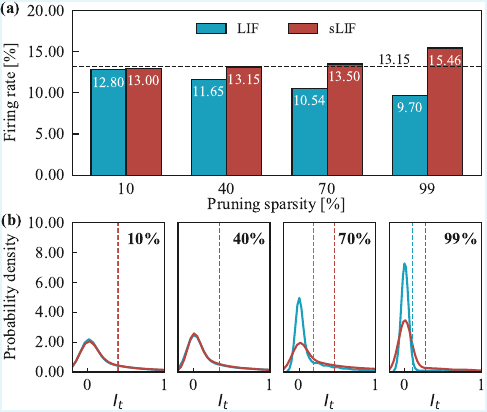}
	\caption{(a) Firing rate comparison of pruned models with LIF- and sLIF-based compensation in the ST modules. The black dotted line indicates the firing rate of the baseline model. (b) Probability density distribution of the input current in $x_{\mathrm{attn}}$ spiking layers.}
	\label{Fig9}
\end{figure}

\subsection{Effectiveness of Pruning Methods}
To enable efficient model pruning, accurately evaluating the importance of individual components within the model is essential. In the proposed L$_{1}$P method, which is designed for constructing sparse weight matrices, the L$_{1}$P norm is employed to quantify the significance of each matrix element. Elements are then pruned based on their ranking derived from the L$_{1}$P importance scores. As illustrated in Fig.~\ref{Fig11}, under a pruning sparsity of $p = 90\%$, only a small subset of elements with the highest L$_{1}$ importance is retained in the sparse weight matrix. To validate the effectiveness of this selection criterion, we conducted a comparative experiment in which elements were pruned randomly (Tab.~\ref{Tab6}). The results show that models pruned using the random strategy exhibit significantly lower classification accuracy compared to those pruned using the L$_{1}$P method. Notably, in the random pruning scenario, a sharp performance drop is observed at a pruning sparsity of $p = 30\%$, whereas the same level of degradation occurs at $p = 70\%$ with the L$_{1}$P approach.

Similarly, in the DSP method, pruning is performed by eliminating dimensions with low significance scores computed using the DVA metric, resulting in a low-rank approximation of the original weight matrix (see Fig.~\ref{Fig11}). To evaluate the effectiveness of this strategy, we also performed a comparative analysis by randomly pruning dimensions. The model performance with random pruning degrades rapidly, with a noticeable accuracy drop occurring before $p = 10\%$, substantially earlier than the degradation point at $p = 30\%$ observed with the DSP method. Moreover, across pruning sparsities of $p \leq 60\%$, models pruned with the DSP strategy consistently outperform those with random pruning. Although the post-pruned accuracy in random pruning is slightly higher than that in DSP, this phenomenon is reasonable given the extremely low accuracy.

In summary, the experimental results validate the effectiveness of the proposed L$_{1}$P and DSP methods, which offer principled importance metrics for unstructured and structured pruning strategies, respectively.

\subsection{Mechanism of Synergistic Learning-Based Compensation}
To evaluate the contribution of synergistic learning to model performance, we visualize the evolution of the IP parameters, $u_{\mathrm{th}}$ and $\tau$, associated with the $Q$, $K$ and $V$ embeddings in the first ST block on the CIFAR10 and CIFAR10-DVS datasets. As illustrated in Fig.~\ref{Fig8}, the magnitude of IP parameter adaptation is positively correlated with the sparsity level. Moreover, the evolution is more pronounced on the more challenging CIFAR10-DVS dataset. This phenomenon is consistent with the theoretical analysis from the perspective of neural dynamics, indicating that IP parameters in synergistic learning play a significant role in enhancing model performance under increased task difficulty. 

\begin{table}[!t]
	\caption{Accuracy comparison of lightweight models compensated with various IP mechanisms.}
	\centering
	\setlength{\tabcolsep}{3.4mm}{
		\begin{tabular}{ccccc}
			\hline
			\multirow{2}{*}{Neuron}   & \multicolumn{2}{c}{$\mathrm{L_{1}P}$} & \multicolumn{2}{c}{DSP} \\
			& $p=95\%$ &          $p=99\%$          & $p=95\%$ &   $p=99\%$   \\ \cline{1-5}
			LIF             & 81.80\%  &          64.60\%           & 83.74\%  &   77.93\%    \\
			PLIF~\cite{Fang_2021_ICCV} & 87.63\%  &          71.96\%           & 89.76\%  &   86.03\%    \\
			STL~\cite{sun2023synapse}  & 87.75\%  &          72.30\%           & 89.79\%  &   86.00\%    \\
			sLIF            & \textbf{88.86\%}  &          \textbf{74.93\%}           & \textbf{90.64\%}  &   \textbf{87.94\%}    \\ \hline
	\end{tabular}}
	\label{Tab_add1}
\end{table}

In SNN models, maintaining an appropriate firing rate is essential for preserving information representation and ensuring stable spike propagation. This requirement is particularly critical in lightweight models, where a moderate increase in firing rate enables a larger population of neurons to participate in encoding, thereby reducing redundancy and mitigating information loss. As illustrated in Fig.~\ref{Fig9}(a), the firing rates of pruned models with LIF-based compensation progressively decrease as pruning sparsity increases. In contrast, models employing sLIF-based compensation consistently restore firing rates to baseline levels. Notably, at an extreme pruning sparsity of $p=99\%$, the firing rate slightly exceeds the baseline, aligning with expectations. Although the elevated firing rate may incur marginally higher power consumption, this overhead remains acceptable given the performance gains achieved by the post-compensated models. Furthermore, we analyze the probability density distribution of input currents in the $x_{\mathrm{attn}}$ spiking layers, generated by the scaled dot-product attention mechanism~(Fig.~\ref{Fig9}(b)). At higher pruning sparsity levels ($p=70\%$ and $p=99\%$), clear differences emerge between LIF- and sLIF-based compensation. The synergistic learning mechanism produces wider distributions with larger expectations, underscoring its role in post-pruning information compensation.

Additionally, we conduct experiments to compensate pruned models using different intrinsic plasticity mechanisms. Prior studies have enhanced neuronal expressiveness by optimizing intrinsic properties, such as the membrane time constant $\tau$~(PLIF)~\cite{Fang_2021_ICCV} or the firing threshold $u_{\text{th}}$~(STL)~\cite{sun2023synapse}. However, compensation based on synergistic learning achieves more superior performance, as reported in Tab.~\ref{Tab_add1}. As task difficulty increases, constructing more detailed neuron models becomes an effective means of improving the performance of SNNs~\cite{CAI2026108417,pnas2513319122,liu2024dendritic}. Compared with these approaches, synergistic learning provides a simple yet efficient alternative, yielding improved training performance without introducing additional computational overhead during the inference stage.

\section{Discussion and Conclusion}\label{Discussion and Conclusion}
In this work, we propose combining synapse pruning and synergistic learning-based information compensation to enhance efficient ST-based models. Specifically, we introduce two synapse pruning strategies—unstructured and structured—tailored to transformer blocks in SNN models to derive compact ST-based architectures. During the fine-tuning phase, synergistic learning is applied to models incorporating sLIF neurons to compensate post-pruning performance. The model size can be flexibly controlled by specifying the desired sparsity level. Experimental results demonstrate that the proposed method effectively compresses various ST-based models. Furthermore, the resulting lightweight models exhibit improved inference performance, underscoring their potential for deployment in edge computing systems.

From a biological perspective, synaptic plasticity regulates information transmission by adjusting the strength of inter-neuronal connections, whereas intrinsic plasticity modifies neuronal properties such as membrane time constants and firing thresholds to control excitability. In the brain, these mechanisms operate synergistically to maintain network stability and robustness. When pruning eliminates a substantial portion of synaptic connections, relying solely on synaptic adjustment often results in reduced firing rates and diminished representational capacity. By jointly adapting intrinsic parameters, synergistic learning preserves population firing rates and increases neuronal heterogeneity, thereby enhancing the diversity of temporal and spatial representations. This enables compressed models to encode information effectively across multiple timescales and feature dimensions, even under high sparsity. Such a mechanism parallels the compensatory strategies observed in biological neural systems under damage or resource constraints, explaining why synergistic learning provides more comprehensive and robust information compensation in pruned models. Consistently, our experimental results demonstrate that synergistic learning facilitates superior performance recovery in heavily pruned networks.

The limitations and future work of the proposed compression strategy are discussed below. Our compression strategy offers flexibility in obtaining lightweight models by allowing manual configuration of the predefined pruning sparsity. However, the same pruning sparsity is applied uniformly across all ST blocks, potentially overlooking the varying contributions of different blocks to the overall model performance. As part of future work, a comprehensive evaluation metric should be developed to enable dynamic adjustment of pruning levels across ST blocks based on their relative importance. Hybrid models that integrate diverse architectural components, such as convolutional blocks, MLP blocks, and transformer blocks, have demonstrated increasing potential in addressing complex tasks. Accordingly, extending our method to support a broader range of architectures may further enhance its applicability and impact across various model designs and domains.

\bibliography{refs}
\bibliographystyle{IEEEtran}

\vfill

\end{document}